\documentclass[11pt, a4paper, logo, copyright]{googledeepmind}

\usepackage[authoryear, sort&compress, round]{natbib}
\paperurl{https://proceedings.mlr.press/v300/kalibhat26a.html}
\usepackage[]{mdframed}

\usepackage{dblfloatfix}

\usepackage{amsmath,amsfonts, amsthm, bm}

\def\eqref#1{equation~\ref{#1}}

\def\1{\bm{1}}

\DeclareMathAlphabet{\mathsfit}{\encodingdefault}{\sfdefault}{m}{sl}
\SetMathAlphabet{\mathsfit}{bold}{\encodingdefault}{\sfdefault}{bx}{n}

\newcommand{\E}{\mathbb{E}}

\newtheorem{defn}{Definition}[section]

\usepackage{listings}
\usepackage{hyperref}
\usepackage{url}
\usepackage{graphicx}
\usepackage{amsmath} %
\usepackage{mathrsfs} %
\usepackage{etoolbox}
\usepackage{cleveref}
\usepackage{tcolorbox}
\usepackage{colortbl}
\usepackage{booktabs}       %
\usepackage{amsfonts}       %
\usepackage{nicefrac}       %
\usepackage{microtype}
\usepackage{enumitem}%
\setlist[itemize]{noitemsep, topsep=0pt}
\usepackage{enumitem,kantlipsum}
\usepackage{xspace}

\usepackage{algorithm}
\usepackage{algpseudocode}
\usepackage{subfig}
\usepackage{subcaption}
\usepackage{tikz}
\usetikzlibrary{arrows.meta, positioning, fit, calc}
\usepackage{tgheros}
\usepackage{multicol}
\usepackage{multirow}
\crefname{section}{\S}{\S\S}
\Crefname{section}{\S}{\S\S}
\Crefname{appendix}{\S}{\S\S}
\crefname{appendix}{\S}{\S\S}

\newlength\savewidth

\definecolor{baselinecolor}{HTML}{d6eaf8}

\definecolor{mygray}{gray}{0.4}

\AtBeginEnvironment{tcolorbox}{\tiny}

\newcount\Comments  %
\usepackage{color}
\definecolor{darkred}{rgb}{0.9,0,0}
\definecolor{darkgreen}{rgb}{0,0.5,0}
\definecolor{darkblue}{rgb}{0,0,0.7}
\definecolor{purple}{rgb}{.6, 0,.6}
\definecolor{orange}{rgb}{1.0,0.64,0}
\newcommand{\kibitz}[2]{\ifnum\Comments=1\textcolor{#1}{#2}\fi}

\title{Interpreting and Controlling Model Behavior via Constitutions for Atomic Concept Edits}
\correspondingauthor{Neha Kalibhat (nehamk@google.com), Zi Wang (wangzi@google.com), Mani Malek (manimalek@google.com). A shorter version has been accepted to the Twenty-Ninth Annual Conference on Artificial Intelligence and Statistics (AISTATS 2026).}

\reportnumber{001} %

\renewcommand{\today}{2026-03}

\author[1]{Neha Kalibhat}
\author[1]{Zi Wang}
\author[1]{Prasoon Bajpai}
\author[1]{Drew Proud}
\author[1]{Wenjun Zeng}
\author[1]{Been Kim}
\author[1]{Mani Malek}

\affil[1]{Google DeepMind}

\begin{abstract}
\vspace{-1em}
We introduce a black-box interpretability framework that learns a verifiable constitution: a natural language summary of how  changes to a prompt affect a model's specific behavior, such as its alignment, correctness, or adherence to constraints. Our method leverages atomic concept edits (ACEs), which are targeted operations that add, remove, or replace an interpretable concept in the input  prompt. By systematically applying ACEs and observing the resulting effects on model behavior across various tasks, our framework learns a causal mapping from edits to predictable outcomes. This learned constitution provides deep, generalizable insights into the model. Empirically, we validate our approach across diverse tasks, including mathematical reasoning and text-to-image alignment, for controlling and understanding model behavior. We found that for text-to-image generation, GPT-Image tends to focus on grammatical adherence, while Imagen 4 prioritizes atmospheric coherence. In mathematical reasoning, distractor variables confuse GPT-5 but leave Gemini 2.5 models and o4-mini largely unaffected. Moreover, our results show that the learned constitutions are highly effective for controlling model behavior, achieving an average of $1.86$ times boost in success rate over methods that do not use constitutions. Our open-sourced code can be found at \url{https://github.com/google-deepmind/atomic_concept_edits}.

\end{abstract}

\begin{document}

\maketitle

\section{INTRODUCTION}

The rapid advancement of large generative models brings with it evolving challenges in safety, reliability, and interpretability. Addressing these issues is paramount as these systems become integral to critical applications. 
A major obstacle is the attribution of models' misbehavior: given the vast scale of modern architectures and data, it is increasingly difficult to trace failures back to specific input features. 
While studies on adversarial machine learning~\citep{jia-liang-2017-adversarial, alzantot2018generatingnaturallanguageadversarial, zou2023universaltransferableadversarialattacks} have shown that subtle input changes can trigger major failures, analyzing these vulnerabilities becomes increasingly difficult in the era of multimodal LLMs~\citep{adversarialmlproblemsgetting}, where the interpretability of prompt perturbations remains a significant research gap.

In this paper, we propose to learn a natural language constitution, describing which categories of \emph{systematic, interpretable} prompt mutations changes the model's behavior. These mutations are designed to target minimal (atomic) modifications that result in significant shifts in model behavior with respect to a task, e.g., text-to-image [mis]alignment, measured by an autorater. The constitution is written in the form of prompt mutation strategies that impact the model's outcome, e.g., ``removing elements already strongly implied or associated with remaining concepts''. 

\begin{figure*}[t]
\centering
\includegraphics[width=0.95\textwidth]{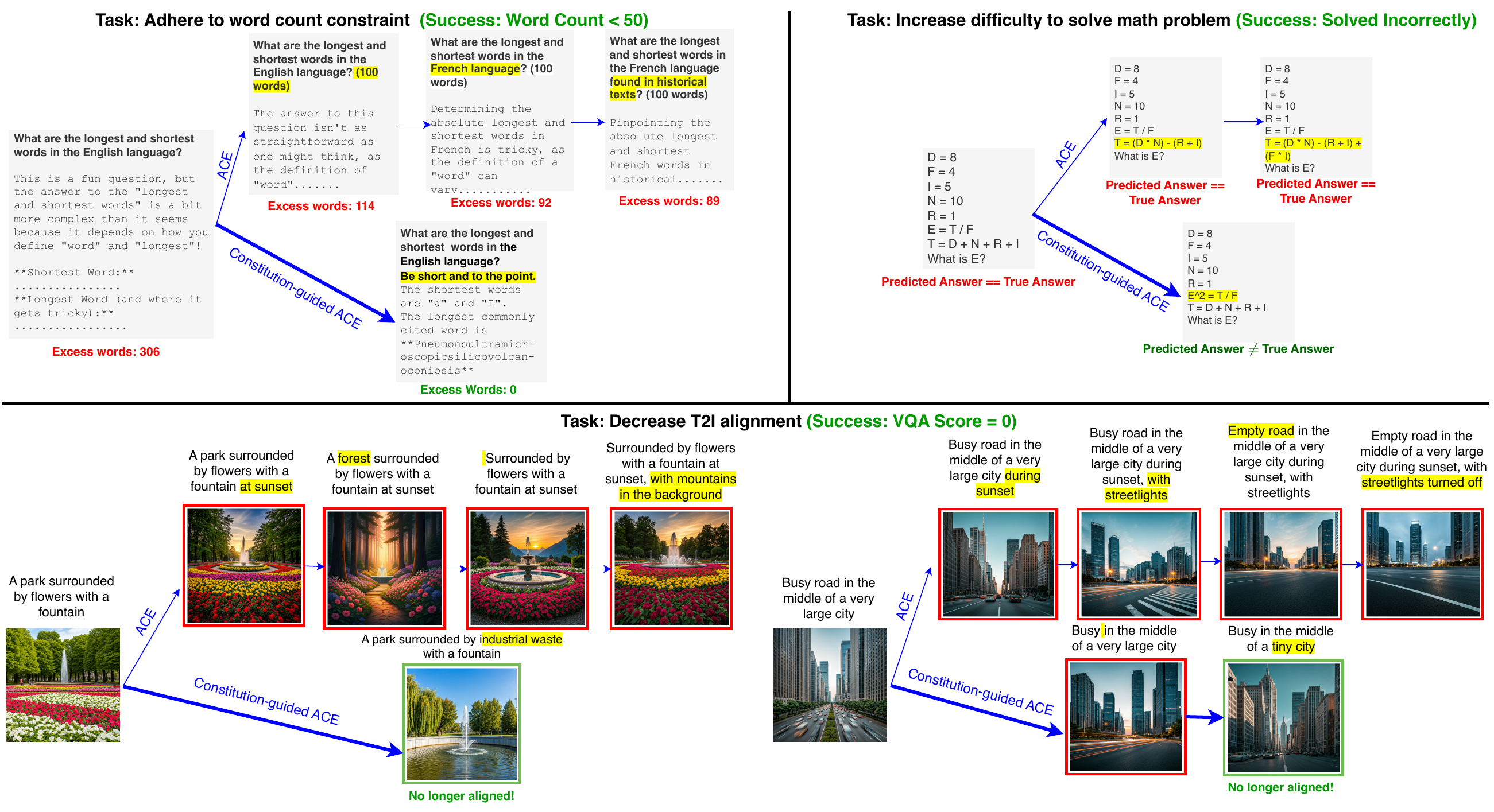}
\caption{\textbf{Qualitative Examples of ACE:} We illustrate how ACE may create mutations that eventually satisfy a given task. In the top-left panel, we demonstrate `Word Count' prompts on Gemini-2.5-Flash with the goal of adherence to a word constraint. In the top-right panel, we demonstrate the `Math' prompts on  GPT o4-mini with the goal of making the problem difficult to solve. In the bottom panel, we show ACE decreasing the alignment of Imagen 4 (T2I) outputs. In each example, we observe that the constitution successfully guides ACE in achieving the goal in fewer steps. More examples shown in Figure \ref{fig:ace_examples}.}
\label{fig:sequences}
\end{figure*}

Our core contribution is a novel framework based on atomic concept edits (ACEs). It is designed to be general-purpose for any target generative model and relies on 3 main inputs: i) a task definition, ii) a dataset of initial input prompts, and iii) a task-specific autorater. The task definition specifies the aspect of a target model we would like to examine by providing a short description of a mutation goal to achieve. For example, to study the alignment of a text-to-image (T2I) model, the task definition can be ``Decrease the alignment between the prompt and the generated image''; to audit the math skill of an LLM,
the goal might be to ``Update the equation to elicit an incorrect answer.''

Using an LLM-based ACE module guided by the task definition, we mutate the initial prompts by applying a series of \textit{atomic} edits given by \emph{add}, \emph{remove}, or \emph{replace} to a single concept in the prompt. For example, given a prompt ``A sheep in a grassy field'', ACEs may include ``Replace sheep with goat'', ``Set the color of the sheep as black'', ``Remove field''. We employ a task-specific autorater (e.g., a text-image alignment evaluator, a math solution verifier) to measure the impact of each ACE. This exploration phase yields interesting findings of the causal effects of ACEs on model behavior, upon which we develop a \textit{constitution}-based sampling approach for ACEs. We use an optimization module to iteratively evolve the constitution using feedback from the explored space of ACEs. The optimized constitution provides an in-depth view of the various modes of ACEs that affect a model's outcome according to the given task. Finally, the ACE generator module can utilize the learned constitution to guide its exploration to successfully steer input prompts towards the goal of the task.

\begin{figure*}[t]
    \centering
    \includegraphics[width=\linewidth]{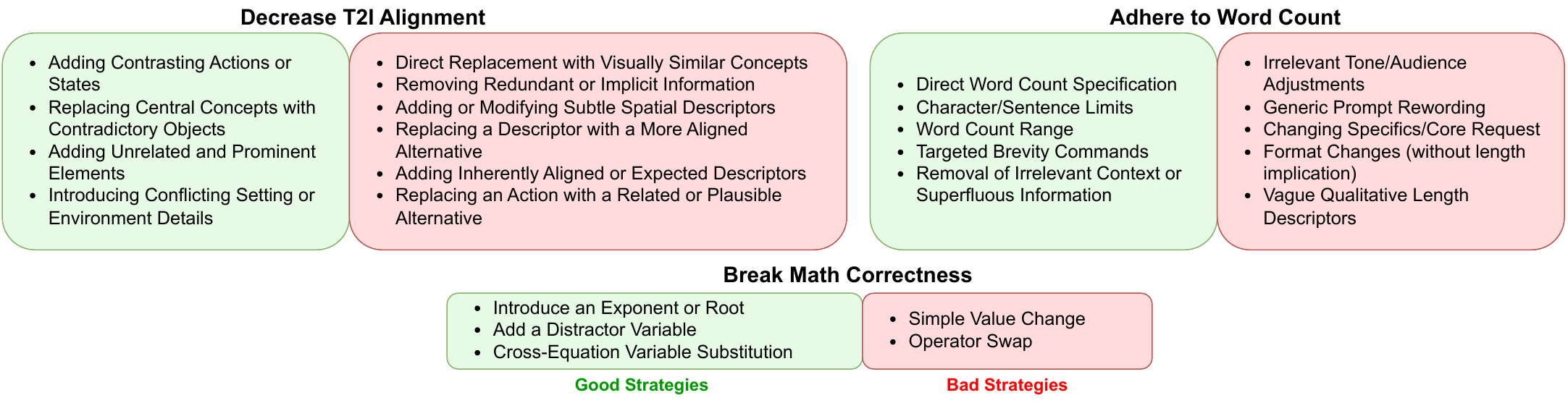}
    \caption{\textbf{Optimized Constitutions for Various Tasks:} We explore ACE patterns to optimize a detailed yet generalizable constitution that both summarizes key aspects of mutations that cause desired model outcomes (defined by the task) and steers unseen prompts towards the same outcome with a minimal number of mutations. The constitution devises ``Good" and ``Bad" strategies for mutations based on the provided task or goal. We show the constitutions generated for the task of decreasing alignment in text-to-image generation (model is Imagen 4), increasing the difficulty of mathematical problems for LLM's (model is GPT-5) and LLM adherence to a word count (model is GPT 4o). More detailed constitutions are shown in the Appendix.}
    \label{fig:learned_constitution}
\end{figure*}

Our framework explores and attributes model behaviors to conceptual patterns in the input prompt space. We demonstrate these findings with qualitative examples of ACEs (See Figure \ref{fig:sequences}) and the learned constitution (See Figure \ref{fig:learned_constitution}) on tasks for decreasing T2I alignment, adhering to word count and breaking math correctness. The same framework also serves as a foundation to help steer model behavior towards any autorater-defined objective. We measure the success rate and diversity of constitution-guided ACEs to confirm the validity of our framework. Our quantitative analyses show that controlling models with learned constitutions achieves $1.86$ times average improvement in success rate over methods that do not use them, while keeping diversity similar for a range of models.

The constitutions also allow for a comparison of model behaviors. For example, GPT-Image's alignment degrades when critical relational and action-focused elements are removed (e.g., ``Removing a Critical Relational Element''), suggesting a strong reliance on strict compositional logic and the prompt's grammatical structure. Conversely, Imagen shows decreased alignment when presented with unrelated or conflicting setting elements, indicating it prioritizes the holistic scene's context and aesthetic coherence. Furthermore, in mathematical reasoning tasks, strategies like introducing distractor variables significantly degrade the performance of GPT-5, a weakness that is notably absent in the Gemini 2.5 Flash, Gemini 2.5 Pro, and o4-mini models.

The primary contributions of this work are:
\begin{itemize}
    \item A generic method for prompt mutation with explainable local edits called ACEs.  
    \item An evolutionary methodology to optimize a constitution that can both explain and steer model behavior, by guiding ACEs towards a given task goal.
    \item We evaluate the effectiveness of ACE, through experiments across three domains with a comprehensive set of metrics. Furthermore, we demonstrate its generalizability across multiple tasks and LLMs.
    
\end{itemize}

\begin{figure*}
    \centering
    \includegraphics[width=\linewidth]{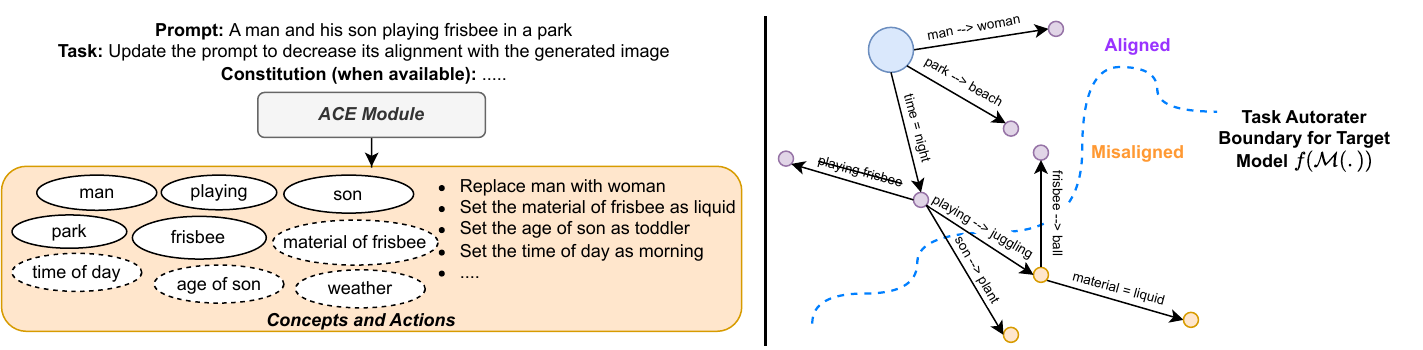}
    \caption{\textbf{ACE Generation:} On the left, we illustrate how the ACE generator module extracts concepts and proposes atomic \textit{add}, \textit{remove} or \textit{replace} ACEs to steer a given prompt towards the goal of satisfying the given task. ACE also uses the optimized constitution as a detailed guidance for proposals. Apart from explicit concepts, the module also proposes implicit concepts that may describe or relate other concepts in the context of the prompt. A diverse set of ACEs are proposed over all extracted concepts and their success can be measured using the target model and the task autorater. ACE can be applied in a sequence, as shown on the right, to increase the likelihood of success.}
    \label{fig:action_generation}
\end{figure*}

\section{BACKGROUND AND RELATED WORK}

\textbf{Interpretability and causality.} Explaining why a model makes a certain prediction is one of the core problems in interpretable ML and causality~\citep{woodward2005making, halpern2005causes, doshi2017towards}. Local interpretability approaches often attribute the model decision to input features (e.g., pixels, tokens) through  perturbation~\citep{chang2018explaining, modarressi2023decompx, miglani2023using} or gradient-based estimation~\citep{sundararajan2017axiomatic, shrikumar2017learning}. To achieve global interpretability, CaCE~\citep{goyal2019explaining} explains model predictions by analyzing the causal effects of including or excluding a concept (e.g., the presence of a car) in the input image. Our approach learns global explanations for language models by analyzing the effects of atomic linguistic concepts, enabling adversarial steering and active testing applications.

\textbf{Adversarial attacks.}
The field of Adversarial ML demonstrates how minimal input changes can
cause dramatic model failures. This vulnerability was first established in computer 
vision \citep{szegedy2014intriguing, goodfellow2015explaining} and later sharpened by 
powerful optimization-based attacks \citep{madry2018towards}. This 
same vulnerability persists in modern systems, where recent work shows how automated red-teaming and simple textual attacks can jailbreak large language models into generating 
harmful content \citep{perez2022red, zou2023universal, wei2023jailbroken}. %
However, these attacks, such as altering a few pixels in an image of a cat, are typically non-interpretable.

\textbf{Active learning and testing.} Our work is also related to active learning for classifiers~\citep{settles.tr09, cohn1996active} with synthesized queries. Similar problems have been studied in relevant fields such as concept learning~\citep{angluin1988queries}, experimental design~\citep{chaloner1995bayesian}, level set estimation~\citep{bryan2005active} and Bayesian optimization~\citep{kushner1962versatile}, whose synthesized queries are typically in a continuous space or finite discrete space. Instead, our queries are in a prompt space. Active testing~\citep{li2024active, kossen2022active, kossen2021active} approaches are similar to active learning, and aims to reduce the evaluation costs by downsampling existing test data. Our approach, which can synthesize difficult test examples, presents new opportunities for active testing.

\textbf{Prompt optimization.}
Our framework learns a constitution for how to mutate a prompt. In prompt optimization, the prompt to be optimized is usually updated by another mutation prompt, which is typically written by hand~\citep{zhou2023large}. 
The constitution optimization step in our approach can make use of existing prompt optimization methods, such as automated instruction generation, using the LLM itself as an optimizer \citep{yang2023large}, forming text-based ``gradients'' as in ProTeGi \citep{pryzant2023automatic}, and tree search in the prompt space with error feedback~\citep{wang2023promptagent}.

\textbf{``Constitutions'' in AI.}
Prior work has used the term ``constitution'' in AI to mean either human-defined safety principles~\citep{bai2022constitutional} or learned user-interaction rules~\citep{petridis2024constitutionmaker}. Here, we use it to describe the principles for how small prompt edits can trigger significant changes in model behavior.

\textbf{Concepts.} 
In the literature of concept learning~\citep{bruner2017study, angluin1988queries, tenenbaum1998bayesian, lombrozo2006structure}, a concept can be viewed as a boolean function, mapping from an object to a whether a concept applies to the object. For example, an object can be a text description, and a concept can be the writing style of the text. 

A recent application of this idea in the T2I domain is the belief graph~\citep{hahn2024proactive}. Belief graphs model the content of an underspecified text prompt as a probability distribution over a constrained set of concepts, which are limited to concrete entities, their attributes, and the relationships between them. This structure allows for highly interpretable edits to the prompt; for example, an image can be systematically modified by adding or removing a specific concept node (e.g., adding the ``wearing a hat'' concept to an ``entity: dog'' node).

We extend the core idea of belief graphs to support both the T2I and the T2T domains. Our work generalizes the framework to accommodate a wider range of generic concepts, moving beyond simple entities and attributes. The primary focus is to leverage this conceptual structure not just for generation, but as an interpretable mechanism for editing text prompts to control model behavior.

\section{PROBLEM FORMUTATION}
\label{sec:problem_formulation}

Our primary goal is to understand how conceptual edits to a prompt alter a model's behavior. To formalize this problem, we first introduce our key terminology: autorater, atomic concept edit (ACE), and constitution. We then present our formal objective in §\ref{ssec:obj}.

\begin{defn}
Given a model $\mathcal{M}: \mathcal X \mapsto \mathcal Z$, an autorater is a binary classifier $f: \mathcal Z \mapsto \{0, 1\}$ that maps a model output $z \in \mathcal Z$ to a desirability score, where $1$ indicates a desirable outcome and $0$ indicates an undesirable one.%
\end{defn}
For example, an autorater for a T2I model could be a visual-question-answering classifier that tests for text-to-image alignment. The task could be ``decrease T2I alignment''. 
Then, the corresponding autorater would assign a score of $0$ to aligned images and $1$ to mis-aligned ones. In this way, the autorater provides a concrete measure of the model's adherence to a specific behavioral rule.

\subsection{Atomic Concept Edits (ACEs)}
Given an input $x\in \mathcal X$, we can obtain a set of existing or explicit concepts $C_x$ in the input $x$ and a set of potential or implicit concepts $C'_x$ that possibly can be included in $x$. For instance, these concepts can be obtained by calling an LLM to parse an input with few-shot examples (more details in \S\ref{sec:method}). Conditioned on the input $x$, a single edit (ACE) to input $x$\footnote{We may view a prompt as a state. Then, ACEs are actions that can be taken at the state.}, can take one of the following forms,

$remove(c)$ -- removing concept $c\in C_x$ from input $x$;

$add(c)$ -- integrating concept $c\in C'_x$ into input $x$;

$replace(c, c')$ -- replacing an existing concept $c \in C_x$ in input $x$ with a different concept $c' \in C'_x$.

For example, \Cref{fig:action_generation} illustrates the concepts and ACEs for an input prompt $x$ related to ``a man and his son playing frisbee in a park''. Explicit concepts $C_x$ include ``man'', ``son'', ``frisbee'', etc. Implicit concepts $C'_x$ include ``time of day'', ``age of son'' etc. An ACE, $replace(\texttt{man}, \texttt{woman})$, can be verbalized as ``replace man with woman''. Once the ACE is applied, input $x$ would have ``a woman and \underline{her} son playing frisbee in a park'', to reflect the concept change, rather than only changing the word ``man''.
 
\subsection{Goal: Learning Constitutions for ACEs}
\label{ssec:obj}
A constitution, denoted $\pi$, is a natural language description of strategies for generating ACEs that modify an input $x$ to elicit more desirable model behavior. Given a task definition $t$ and a constitution $\pi$, we can define a distribution over the set of all possible ACEs: $p(a \mid x, t, \pi)$ for any input $x$.

Our goal is to learn a constitution $\pi$ that generates ACEs which improve the likelihood of achieving desirable behavior (satisfying task $t$), as judged by the autorater $f$. Let $\phi(x, a)$ be the function that applies an ACE $a$ to an input $x$ to produce a transformed input. Formally, our objective is to
\begin{align}
\label{eq:obj}
    \underset{\pi}{\text{max}} \,\,\, \E_x\left[\E_{a\sim p(a \mid x, t, \pi)}\left[f(\mathcal M(\phi(x, a))) \right]\right].
\end{align}
This objective implicitly assumes a distribution over model input, which is a common assumption in machine learning. By taking an expectation over ACEs, \Cref{eq:obj} essentially optimizes the probability that the constitution-guided ACEs can lead to better model behavior.  
Achieving this objective presents two main challenges: (1) how to obtain the ACE distribution $p(a \mid x, t, \pi)$, and (2) how to perform the optimization in \Cref{eq:obj}.

\begin{figure}
    \centering
    \includegraphics[width=0.98\linewidth]{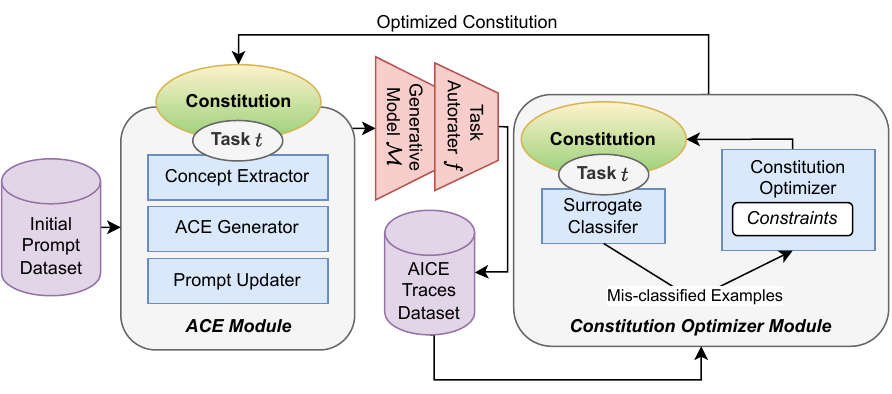}
    \caption{\textbf{ACE Framework with Constitution Optimization:} Our framework consists of an ACE Module - which converts a set of initial prompts into a dataset of ACE traces labeled by the target model and an autorater. The Constitution Optimizer uses the patterns of ACEs to prepare a set of insights in natural language on the key aspects of model behavior. %
    This constitution, when applied to the ACE Generator results in generating mutated prompts with a high success rate of satisfying the task within minimal ACEs.}
    \label{fig:framework}
\end{figure}

\section{OUR METHODS}
\label{sec:method}
This section details our approach. For solving \Cref{eq:obj}, we propose a practical method for generating ACEs (§\ref{ssec:cace}) and optimize the constitution (§\ref{ssec:constitution_optimization}). We explain how a learned constitution can be used to control model behavior at inference time (§\ref{ssec:control}). To ensure clarity, we call $\mathcal M$ the target model in the following sections.

\subsection{Constitution-guided ACE Module}
\label{ssec:cace}
We use constitution to guide the discovery of ACEs following a 3-stage process, involving Concept Extraction, ACE Generation and Prompt Mutation. Given a prompt $x$, a task definition $t$ and a constitution $\pi$ (when available, otherwise is empty), the Concept Extractor distills the prompt into a set of 
concepts. The ACE Generator proposes ACE's for each concept and the Prompt Mutator updates the prompt. 

This process when applied to a number of initial prompts, yields a dataset of ACEs to steer prompts with ground truth scores measuring the success with respect to the given task. The ACE dataset for $N$ initial prompts is represented as $D = \{(x_i, a_j, y_j)\}_{i=1,j=1}^{N, M}$ where $y_i = f(\mathcal M(\phi(x_i,a_i)))$ represents the ground-truth autorater scores of $x_i$, updated with ACE $a_i$.%
We explicitly instruct the Concept Extractor and ACE Generator to explore various possibilities and to promote diversity to increase the likelihood of traversing interesting, non-trivial insights into model behavior.

\subsection{Constitution Optimization Module}
\label{ssec:constitution_optimization}
We use automatic prompt optimization to optimize the constitution. \Cref{alg:constitution_optimization_alg} presents the algorithm.

\Cref{eq:obj} uses feedback from target model $\mathcal{M}$ and autorater $f$ to optimize the constitution $\pi$ on the set of sampled mutations over all prompts $x$. In practice however, target model $\mathcal{M}$ and autorater $f$ are large AI models with highly expensive inference costs. The set of updated prompts after applying sampled ACEs on the prompt dataset can also be large. %
These costs are even more amplified when $\mathcal{M}$ is an image/video generation model. Given that our framework focuses on performing ACE on textual inputs, we leverage the instruction-following power of LLMs to circumvent this issue. 

We therefore use $S$, an instruction template to an LLM that takes task definition $t$, examples from ACE dataset $D$ and constitution $\pi$ as input to predict autorater score $y_i$. We term this LLM, along with its instruction template $S$, as the \textit{surrogate classifier}. The new formulation is as follows:
\begin{align}
    \mathcal{\pi}^* \approx \arg \max_{\pi} \sum_{d \in D}\left[g(LLM(S(\pi, t, d)))\right]
\end{align}

where $g(.)$ is a metric representing the classification accuracy of the surrogate model. We ensure that the examples in $D$ are balanced by their label in practice. Motivated by the success of Alpha Evolve \citep{alphaevolvecodingagentscientific} and TextGrad \citep{yuksekgonul2024textgradautomaticdifferentiationtext}, we use an evolutionary algorithm to iteratively update $\pi$ using feedback of mis-classifications from the surrogate classifier. This feedback serves as a \textit{loss} to optimize the constitution. This method of evolutionary training using a surrogate model approximates the complex and expensive model and autorater, allowing us to rapidly iterate and guide constitution learning.

We design the constitution to be a list of strategies to accomplish the task with 2 sections: Good Strategies and Bad Strategies. Each strategy is a unique method of steering prompts and unlocks a novel aspect of model behavior. The strategies should not contradict or overlap with each other. We also fix the size of the constitution to be $K$ (a hyperparameter we can control) representing the total number of strategies (good and bad included). This plan for designing the constitution is passed to the LLM as a comprehensive set of instructions to help design a generalizable constitution.

\subsection{Using Constitutions for Model Behavior Control}
\label{ssec:control}
Because the constitution can be used for any input, we can apply ACEs iteratively on input $x$ to gain higher probability of success; i.e., getting better model behavior judged by the autorater. In practice, we can get new inputs $n$ ACE steps away from original input $x$, by applying ACE $a_1$ to obtain updated input $x' = \phi(x, a_1)$, applying ACE $a_2$ to obtain $x'' = \phi(x', a_2)$, so on and so forth until the $n$-th ACE.

Applying a sequence of ACEs enables greater exploration of the input space. If there exists non-zero probability of success at each step, we can show that the probability of success approaches 1 as the number of ACE steps goes to infinity.

\begin{algorithm}
\caption{Constitution Optimization Algorithm}
\label{alg:constitution_optimization_alg}

\small
\begin{algorithmic}[1]
\State \textbf{Input:} Training set $D_{train}$, Validation set $D_{val}$, Test set $D_{test}$, Language Model $LLM$, Number of epochs $E$, task definition $t$ \\
\textit{SC} (Surrogate Classifier) is the LLM along with the instruction template $S$ to predict the autorater score for each given example. \\
\textit{CO} (Constitution Optimizer) is the LLM along with an instruction template to update a given constitution using the classification outcome $res$ of the \textit{SC} 
\State \textbf{Initialize:}
\State Generate an initial constitution $\pi^*$ using the $LLM$ and $D_{train}$
\State Set candidate list $\mathcal{L}_{candidates} \gets \{\pi^*\}$

\For{epoch $e = 1$ to $E$}
    \For{ $\pi_i$ in $\mathcal{L}_{candidates}$}
        \State $res_{val}^{(i)} = SC(LLM, \pi_{i}, t, D_{val})$
        \State $acc_{val}^{(i)} = g(res_{val}^{(i)})$ \Comment{$g(.)$ measures the classification accuracy using predictions in $res$}
    \EndFor
    \State $i^{*} = \arg\max_{i} acc_{val}^{(i)}$ \Comment{Find the best candidate using $D_{val}$}
    \State $\pi^* \gets \pi_{i^{*}}$

    \State $res_{train} \gets SC(LLM, \pi^*, D_{train})$
    \State $res_{test} \gets SC(LLM, \pi^*, D_{test})$ \\
    
    \State Shuffle $D_{train}$
    \For{batch $b_j = b_1$ to $b_B$ in $D_{train}$} \Comment{Optimize on batches of $D_{train}$}
        \State $\pi_{j} \gets CO(LLM, \pi^*, b_j, res_{train,j})$ 
    \EndFor
    \State $\mathcal{L}_{candidates} \gets \{\pi_1, \dots, \pi_B\}$
\EndFor
\end{algorithmic}
\end{algorithm}

\begin{table*}

\centering

\caption{\textbf{Measuring the Impact of ACE and Constitution Optimization:} Using 3 diverse tasks, we measure the performance of various generative models (text-to-image and text-to-text) against 3 metrics. We compare the accuracy of the surrogate classifier using an un-optimized constitution (Epoch 0) with an optimized constitution (Epoch E). The  success rate measures the percentage of initial prompts that were successfully steered towards satisfying the task within a maximum ACE sequence length of 4. We also compute the relative increase ($r_1/r_0-1$, where $r_1$ is the success rate for ACE with constitution and $r_0$ is the baseline ACE) in success rate noted in $(\uparrow \cdot)$. We use Self-BLEU to verify if ACE maintains the diversity of prompts after mutations. Across every task and target model tested, using optimized constitutions to modify prompts achieves a 1.86 times average relative increase in  success rate over the baseline.}

\resizebox{\textwidth}{!}{
\begin{tabular}{c|c|c|c|c|c|c|c|c}
\toprule
& & \multicolumn{2}{c|}{\textit{Surrogate Accuracy} $\uparrow$} & \multicolumn{2}{c|}{\textit{Success Rate} $\uparrow$} & \multicolumn{3}{|c}{\textit{Self-BLEU} $\downarrow$} \\
\cline{3-9}
\textbf{Target Model} & \textbf{Task} & \textbf{Initial} & \textbf{Final} & \multirow{2}{*}{\textbf{ACE}} & \textbf{ACE} & \textbf{Starter} & \multirow{2}{*}{\textbf{ACE}} & \textbf{ACE} \\
& & \textbf{Constitution} & \textbf{Constitution} & & \textbf{Constitution} & \textbf{Prompts} &  &  \textbf{Constitution} \\
\midrule
\midrule

Gemini 2.5 Flash & Word Count & 87.46 & 87.46 & $33.32 \pm 3.55$ & \textbf{63.38 $\pm$ 3.30} ($\uparrow$ 90\%) & 0.04 & 0.03 $\pm$ 0.02 & 0.15 $\pm$ 0.02 \\
Gemini 2.5 Pro & Word Count & 94.70 & 94.70 & 25.50 $\pm$ 3.37 & \textbf{74.70 $\pm$ 3.65} ($\uparrow$ 193\%) & 0.04 & 0.03 $\pm$ 0.02 & 0.18 $\pm$ 0.01 \\
GPT 5 &  Word Count  & 93.85 & \textbf{96.08} & 32.67 $\pm$ 3.26 & \textbf{86.02 $\pm$ 2.50} ($\uparrow$ 163\%)  & 0.04 & 0.04 $\pm$ 0.02 & 0.19 $\pm$ 0.01 \\
GPT 5-mini &  Word Count  & 97.49 & 97.49 & 29.74 $\pm$ 3.68 & \textbf{93.47 $\pm$ 2.41} ($\uparrow$ 214\%) & 0.04 & 0.03 $\pm$ 0.02 & 0.31 $\pm$ 0.01 \\
GPT 5-nano &  Word Count  & 93.30 & \textbf{95.81} & $31.37 \pm 3.28$ & 
\textbf{87.11 $\pm$ 2.69} ($\uparrow$ 177\%) & 0.04 & 0.03 $\pm$ 0.02 & 0.35 $\pm$ 0.01 \\
GPT 4o &  Word Count  & 93.29 & \textbf{95.81} & 29.92 $\pm$ 3.60 & \textbf{49.24 $\pm$ 4.38} ($\uparrow$ 65\%) & 0.04 & 0.02 $\pm$ 0.01 & 0.02 $\pm$ 0.01 \\
o4-mini &  Word Count  & 91.92 & \textbf{92.47} & 33.84 $\pm$ 3.40 & \textbf{67.54 $\pm$ 4.00} ($\uparrow$ 100\%) & 0.04 & 0.03 $\pm$ 0.02 & 0.09 $\pm$ 0.02 \\
\midrule
Gemini 2.5 Flash & Math & 68.23 & \textbf{78.45} & 17.46 $\pm$ 4.92 & \textbf{46.62 $\pm$ 7.98} ($\uparrow$ 167\%) & 0.35 & 0.07 $\pm$ 0.06 & 0.10 $\pm$ 0.02 \\
Gemini 2.5 Pro & Math & 62.29 & \textbf{66.92} & 22.53 $\pm$ 5.83 & \textbf{82.33 $\pm$ 4.95} ($\uparrow$ 265\%) & 0.35 & 0.07 $\pm$ 0.05 & 0.24 $\pm$ 0.01 \\
GPT 5 & Math & 68.76 & \textbf{76.20} & 13.40 $\pm$ 4.74 & \textbf{61.82 $\pm$ 8.28} ($\uparrow$ 361\%) & 0.35 & 0.02 $\pm$ 0.03 & 0.23 $\pm$ 0.01\\
o4-mini & Math & 55.40 & \textbf{69.45} & 10.07 $\pm$ 4.83 & \textbf{87.40 $\pm$ 5.09} ($\uparrow$ 768\%) & 0.35 & 0.01 $\pm$ 0.03 & 0.17 $\pm$ 0.04 \\
\midrule
Imagen 4 & T2I Align & 82.14 & \textbf{92.08} & 83.61 $\pm$ 5.12 & \textbf{97.00 $\pm$ 2.72} ($\uparrow$ 16\%) &  0.04 & 0.03 $\pm$ 0.01 & 0.03 $\pm$ 0.01 \\
DALL-E 3 & T2I Align & 79.82 & \textbf{91.34} & 70.83 $\pm$ 5.93  & \textbf{87.13 $\pm$ 4.30} ($\uparrow$ 23\%) & 0.04 & 0.02 $\pm$ 0.02 & 0.01 $\pm$ 0.01 \\
GPT-Image & T2I Align & 82.47 & \textbf{91.25} & 91.06 $\pm$ 4.20 & \textbf{98.97 $\pm$ 1.02} ($\uparrow$ 9\%) & 0.04 & 0.03 $\pm$ 0.01 & 0.03 $\pm$ 0.00 \\
\bottomrule
\end{tabular}
}

\label{tab:model_results}
\end{table*}

\section{EXPERIMENTS}

In this section, we use ACE and Constitution-guided ACE (CACE) to understand and steer LLMs and text-to-image (T2I) models on 3 different downstream tasks described in \Cref{ssec:downstream}. We show the qualitative and quantitative results on the learned constitutions, compared with ACE (without constitution) as a baseline. Across our experiments, Gemini 2.5 Flash \citep{geminiteam2024gemini15unlockingmultimodal} was used as the ACE Generator and for Constitution Optimization, and target models include Gemini 2.5 Flash and Pro~\citep{geminiteam2024gemini15unlockingmultimodal}, GPT-5, 5-mini, 5-nano, 4o and OpenAI o4-mini~\citep{openai2024gpt4technicalreport}. See \Cref{sec:exp_appendix} for more details.

\subsection{Downstream tasks and metrics}
\label{ssec:downstream}
Each downstream task has a high-level task description, a dataset of prompts and an autorater that determines whether a model response is desirable. The metrics for all tasks include the surrogate accuracy (i.e., the accuracy of autorater scores predicted by the surrogate), the ACE success rate (i.e., whether the desired model behavior can be achieved by taking ACEs), and the Self-BLEU score \citep{papineni-etal-2002-bleu} for measuring the diversity of a set of prompts. A higher Self-BLEU score implies more repetition in the generated text and therefore less diversity.

Three tasks we tested are below.

\textbf{Word Count}: %
The high level description is ``force the model to adhere to a word count constraint''. The autorater counts the number of words in the model response, and classify whether there are fewer than 50 words (this exact constraint description is not revealed to ACE).  For generating ACEs and learning a constitution, we sample 100 questions from the wikiHow dataset of LIMA~\citep{zhou2023lima} as the initial prompts, such as ``How to cook an omelette?''. We then evaluate the constitution on 100 held-out questions in LIMA.

 \textbf{Math}: %
 The task is to steer model to 
 incorrectly answer math questions.
 We sample and 30 test questions as initial prompts from the GSME dataset in~\citep{li2025questbench}, where each math problem is presented as a constraint satisfaction problem (CSP), such as ``a=1, b=2, c = a+b. What is c?'' The high-level task description is ``Modify the math problem to make it more difficult to solve correctly while ensuring it remains a valid CSP, solvable by a symbolic math solver.'' The autorater uses SymPy~\citep{10.7717/peerj-cs.103} to parse the CSP and solve the equations to obtain the ground truth answer. 

\textbf{T2I Align}: %
The high-level description is ``Update the prompt to decrease its alignment with its generated image''. We use 150 starter text-to-image captions from the COCO dataset~\citep{lin2015microsoftcococommonobjects} as initial prompts. We evaluate three target T2I models: Imagen 4~\citep{saharia2022photorealistictexttoimagediffusionmodels}, DALL-E 3~\citep{ramesh2021zeroshottexttoimagegeneration}, and GPT-Image 1~\citep{openai2024gpt4technicalreport}. For any given prompt, we generate 3 images and use the VQA Score as judged by Gemini~\citep{comanici2025gemini25pushingfrontier} as the autorater, selecting the lowest of the 3 scores as the overall score for the prompt.

\begin{figure*}
    \centering
    \includegraphics[width=\linewidth]{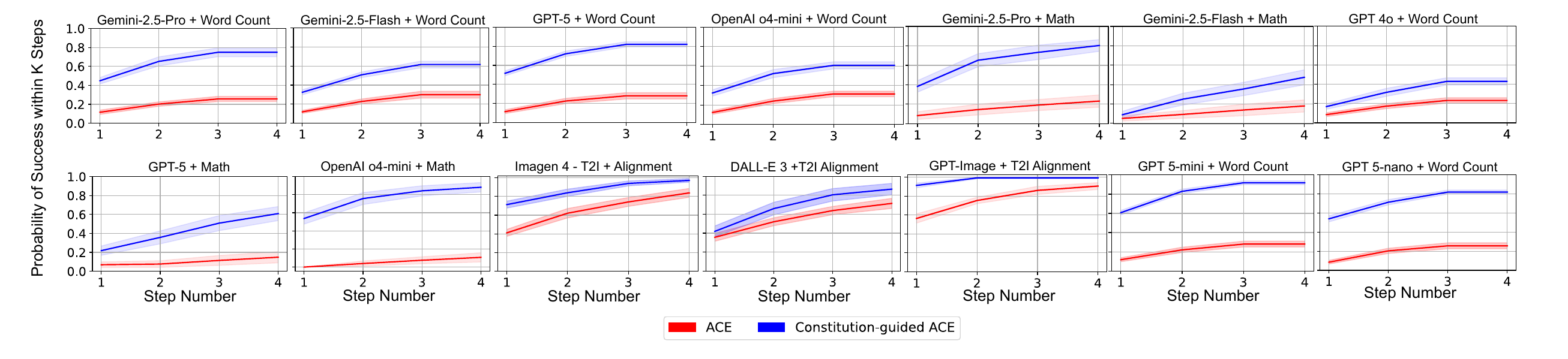}
    \caption{\textbf{Probability of ACEs Succeeding:} We measure the probability of ACE satisfying the goal task within the given number of steps (or length of ACE sequences), represented in the x-axes. We observe that using an optimized constitution with ACE helps achieve the goal faster (blue curves). Within 4 steps, we observe that ACE achieves a high success rate ($>$ 0.8\%) across all models and tasks.}
    \label{fig:success_distribution}
\end{figure*}

\subsection{Qualitative Results: Understanding Models}

For Word Count, since the task is very simple, we know the constitution should include concepts such as limiting the word count to a certain number, being concise etc. The constitution learned for this task is shown in \Cref{fig:learned_constitution}, which verifies that ACE is indeed capable of discovering the right kind of insights, even with very limited information of the task and mutation utilities. In the example illustrated in Figure \ref{fig:sequences}, we observe that baseline ACEs contain long responses to the input until the third step in the sequence which gets a lower score due to a word constraint. In constitution-guided ACE however, the guidance provided helps mutate the prompt to quickly satisfy the word count limit within the first step.    

\Cref{fig:learned_constitution} shows the optimized constitution for T2I Align, Math and Word count tasks. Based on the task and target model, the optimized constitution summarizes unique aspects of conceptual perturbations and groups them as good or bad strategies. We observe that each insight on model behavior is described in a generalizable manner and does not contain any specific examples from the data. It is also important to note that these are actionable strategies that discover non-trivial dimensions and can directly be used to prepare relevant data and debug models. We observe that text-to-image models can struggle with alignment when objects, actions or the environment is either unrealistic or contradictory. This is illustrated in Figure \ref{fig:sequences} where constitution-guided ACEs like ``adding industrial waste'' to a scene with a park and ``replacing very large with tiny'' for a busy city, lead to successfully decreasing the text-to-image alignment. In math tasks, we observe that models sometimes cannot solve equations when complex operations like exponents are added (See Figure \ref{fig:sequences}). Such mutations make the problem more complex, exposing LLM difficulty in solving involved mathematical problems. 

We also observe unique behavior across models (See Appendix \ref{sec:more_constitutions}) and discover non-trivial strategies. For word count constitutions, we do not explicitly specify the word limit of 50 in the task but we discover several successful strategies that indirectly achieve the task instead of directly specifying a word limit. These include “Explicit single-unit constraint”, “Structural constraints”, “Implicit tone modification". 

For T2I generation, GPT-Image shows degrading T2I alignment by removing relational and action-focused elements like "Removing a Critical Relational Element" and "Removing the Object of a Verb". This points to a highly structured semantic understanding of the prompt where the relationships and actions are treated as distinct, critical components. On the other hand, Imagen shows decreased alignment for strategies like "Adding Unrelated and Prominent Elements" and "Introducing Conflicting Setting or Environment Details" which may indicate the model’s on the holistic scene’s context. 

\Cref{fig:math-constitutions} compares the constitutions composed of "attack" strategies to make math equations more difficult for each model. It shows introducing distractor variables may cause GPT-5 to fail at solving simple math problems, but does not tend to impact the correctness of the answers from Gemini 2.5 Flash, Gemini 2.5 Pro and o4-mini. Moreover, shared strategies like introducing exponents demonstrate that LLMs in general still cannot master complex algebraic problems.

The above examples demonstrate that the insights from the constitutions are informative and non-trivial. Moreover, when we use constitution-guided ACE for that task and model, we observe that success is achieved in fewer steps.

\subsection{Quantitative Results: Steering Models}

\paragraph{Constitution Impact and Generality}
We measure the accuracy of surrogate classifier in predicting whether a proposed ACE to a prompt satisfies the given task according to the model and autorater, for the initial constitution (at epoch 0 in the evolution) and the final optimized constitution (final epoch when the evolution ends). Our results in Table \ref{tab:model_results} confirm that the optimization is successful observing improved performance and generalization (on the test split) over training iterations. Depending on the simplicity of the task, like some models assessed on word count, we also observe that the constitution may not evolve significantly over iterations, maintaining the surrogate accuracy. In the Appendix Section \ref{sec:textgrad}, we compare our surrogate classifier to TextGrad \cite{yuksekgonul2024textgradautomaticdifferentiationtext} and observe that we outperform TextGrad on all tasks.

\paragraph{Effectiveness and Success}
Taking a set of unseen initial prompts, we use the ACE generation (with and without constitution-guidance) to randomly generate sequences of ACEs that result in successfully satisfying the task. For every initial prompt, we randomly select one sequence and we measure an aggregate score of the percentage of sequences that were successful, i.e., the ACE success rate. This score is averaged over $10$ seeds and reported in Table \ref{tab:model_results}. We observe that the success rate of ACE is higher with constitution-guidance. In Figure \ref{fig:success_distribution}, we measure the cumulative probability distribution of ACE sequence lengths for satisfying the given task. Specifically, this measures if a successful mutation is achieved for the set of initial prompts within N steps of sampling ACEs (x-axis in Figure \ref{fig:success_distribution}. Using the constitution (blue curves) results in higher likelihood of success in all models and tasks. This further confirms the effectiveness of using an optimized constitution with ACE since it not only achieves success quickly (as shown in Figure \ref{fig:sequences}), but also shows a higher likelihood of success over multiple sequences, as shown in Figure \ref{fig:success_distribution}.

We measure the diversity of initial prompts and the diversity of final prompts to ensure our approach is not simply converging on a trivial solution of similar, less-diverse prompts. The Self-BLEU scores in Table \ref{tab:model_results} confirms this  by showing that the diversity does not significantly decrease after ACEs are applied.

\section{CONCLUSIONS}

We introduced a novel framework, Constitution-guided ACE, for understanding and steering black-box large generative models. The core objective was to systematically decode the ``black box'' behavior of these models by establishing a link between concept-level prompt modifications and shifts in model behavior. Our approach efficiently learns a human-interpretable constitution, which acts as a rulebook for tasks like adversarial steering. 

A key direction for future work is extending this framework to other input modalities and applications such as image, video or agentic systems where, concepts and ACEs may be complex and extracting natural language explanations can be challenging. An added hurdle in such advanced systems can be the unavailability of robust autoraters. We also study model behavior in conjunction with the autorater by assuming the reliability of the autorater which may not be the case for all models and tasks in practice. More research is needed to be able to use ACE to disentangle vulnerabilities in model behavior from those in autorater behavior. We demonstrate applying ACE in sequences (multiple steps, or trajectories) however, the guiding constitution is optimized on single ACEs in turn promoting a greedy approach. Further research is needed to understand if agents are capable of optimally traversing trajectories of ACEs to achieve even higher likelihood of success. 

\section*{ACKNOWLEDGMENTS}
We would like to thank William Isaac and Helen King for valuable discussions and support. We would also like to thank Uri Shalit, Utsav Prabhu, Ken Burke, Caroline Pantofaru, Oyvind Tafjord, John Hewitt, Robert Geirhos for insightful comments and help, Dirichi Ike-Njoku for code review, and anonymous reviewers for feedback on the paper.

\bibliography{references}

\appendix
\newpage
\renewcommand\thefigure{\thesection.\arabic{figure}}
\renewcommand\thetable{\thesection.\arabic{table}}

\setcounter{figure}{0} 
\setcounter{table}{0} 

\section{Additional Experiment Details}
\label{sec:exp_appendix}

We summarize our choices for data, autorater and constitution hyper-parameters in Table \ref{tab:experimental_details}. For ACE generation, we use a fixed number of initial prompts and sample 2 ACEs per prompt, repeating this for each mutated prompt up to a max depth (sequence length). We stop sampling once we reach a prompt that successfully satisfies the task, hit the max depth or encounter an error during concept extraction, mutation, autorater or target model. We therefore report an approximate number of valid ACEs we generated that were used for optimizing the constitution and reporting metrics as this differs slightly among the different target models we studied. 

\begin{figure}[h]
    \centering
    \subfloat{\includegraphics[width=0.33\linewidth]{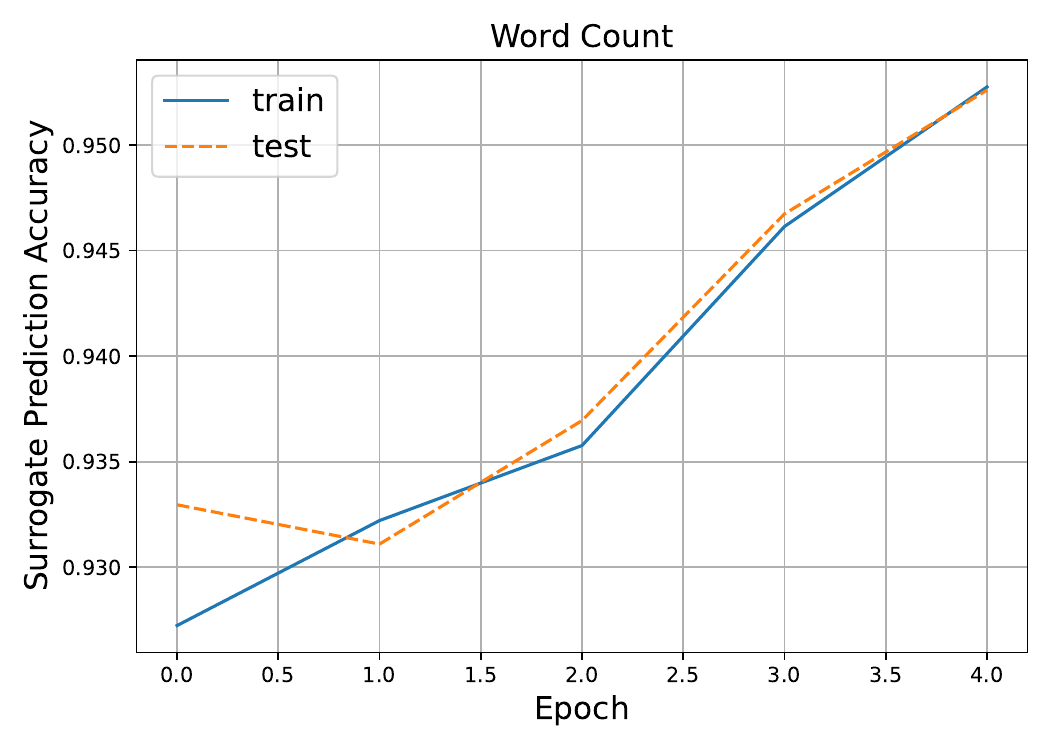}}
    \subfloat{\includegraphics[width=0.33\linewidth]{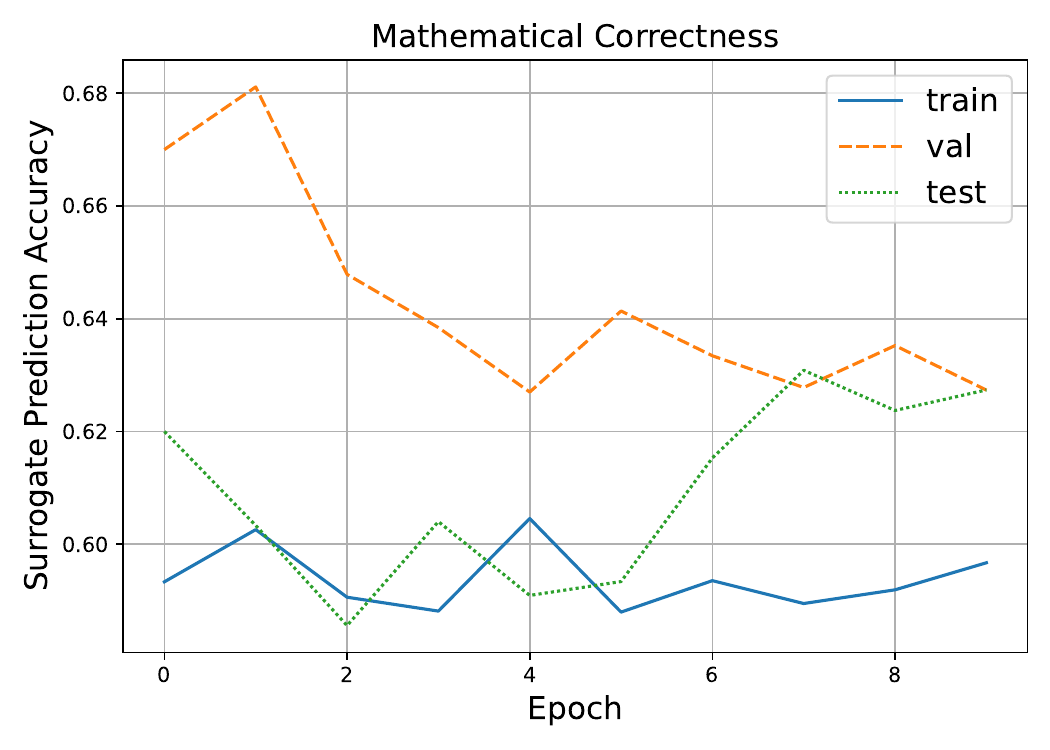}}
    \subfloat{\includegraphics[width=0.33\linewidth]{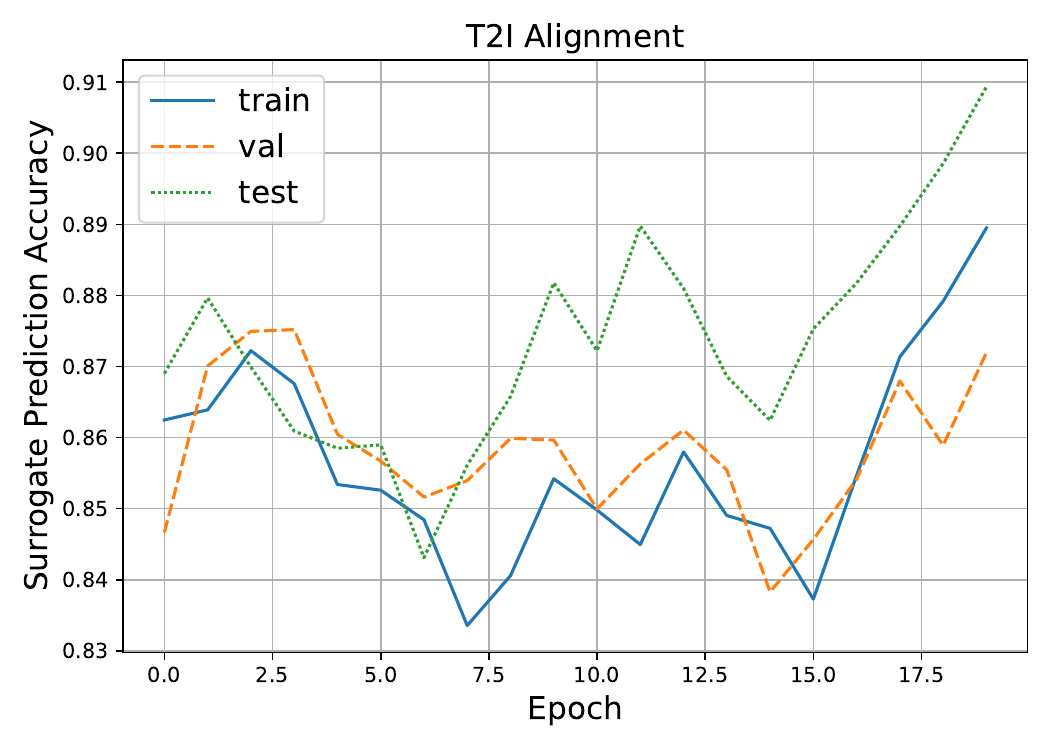}}
    \caption{\textbf{Accuracy of Surrogate Model against Training Iterations}: We plot the classification accuracy of the train, test and validation split over iterations of constitution evolution. We observe that the test accuracy improves over iterations. Simpler tasks require fewer iterations while more complex tasks can benefit from longer training.}
    \label{fig:constitution_training}
\end{figure}

We also experiment with dynamic constitution update strategies. Particularly, per epoch, we control the percentage of strategies to be updated via a hyperparameter. Moreover, we also control the number of strategies to be added or removed in the constitution via another hyperparameter. Both these hyperparameters lead to a dynamic change in both these parameters across epochs. Empirically, for complex tasks like math and T2I, we observed that gradually increasing the number of strategies across epochs, combined with decreasing the allowed percentage change gives the best results as per the performance and stability of training. The accuracy plots over iterations of evolving the constitution are displayed in Figure \ref{fig:constitution_training}.

\begin{table}[h]
\centering
\resizebox{\textwidth}{!}{
\begin{tabular}{c|c|c|c|c}
\toprule
& \textbf{Parameter} & \textbf{Word Count} & \textbf{Math} & \textbf{T2I Align} \\   
\midrule
\midrule
\multirow{3}{*}{\textbf{Task definition}} & \multirow{3}{*}{Task definition} & Force the model to & Modify the math problem & Update the prompt  \\ 
& & adhere to a word & to make it more difficult & to decrease its alignment \\ 
& & count constraint & to solve correctly & with its generated image \\ 
\midrule
\multirow{3}{*}{\textbf{Autorater Details}} & Autorater & Response length & SymPy & Gemini 2.5 Flash \\
& Autorater success & Response length $<$ 50 & Answer $\neq$ Ground truth & VQA Score $=$ 0 \\
& Autorater failure & Response length $\ge$ 50 & Answer $=$ Ground truth & VQA Score $=$ 1 \\
\midrule
 & Initial prompt dataset & LIMA & GSME (QuestBench) & COCO Captions \\
& \# initial prompts & 100 & 100 & 150 \\
\textbf{ACE Generation} & \# ACEs sampled for each prompt & 2 & 2 & 2 \\
& ACEs sequence length & 3 & 3 & 4 \\
& \# valid ACEs generated & $\sim$2500 & $\sim$400 & $\sim$1200 \\
\midrule
& Epochs & 5 & 10 & 20 \\
& Batch Size & 100 & 200 & 50 \\
\textbf{Constitution} & Constitution Size (Epoch 0) & 10 & 5 & 10 \\
\textbf{Optimization} & Constitution Size (Epoch E) & 10 & 10 & 10 \\
& Change \% (Epoch 0) & 10 & 100 & 100 \\
& Change \% (Epoch E) & 10 & 5 & 5 \\
\bottomrule
    \end{tabular}
    }
    \caption{\textbf{Experiment Details:} We summarize all the parameters used for each task for ACE generation and constitution optimization including autorater details, datasets, sizes and hyper-parameters.}
    \label{tab:experimental_details}
\end{table}

\section{Using TextGrad as a Surrogate Classifier}
\label{sec:textgrad}

\begin{table}
    \centering
    \begin{tabular}{c|c|c|c}
    \toprule
\textbf{Task} & \textbf{Model} & \textbf{TextGrad} & \textbf{Surrogate Model (ours)} \\
\midrule
\midrule
Decrease T2I Alignment & Imagen 4 & 76.67 & \textbf{92.08} \\
Adhere to Word Count & GPT-4o & 89.97 & \textbf{95.81} \\
Make Math Incorrect & GPT-5 & 66.66 & \textbf{76.20} \\
\bottomrule
    \end{tabular}
    \caption{\textbf{Using TextGrad as a Surrogate Model:} We measure the accuracy on the test split while using TextGrad for surrogate classification on each task. We observe that our model out-performs TextGrad and results in a more interpretable constitution to understand model behavior.}
    \label{tab:textgrad}
\end{table}

We include results from using TextGrad \cite{yuksekgonul2024textgradautomaticdifferentiationtext} (a SOTA or near-SOTA approach for obtaining textual gradients) as a baseline for surrogate classification i.e., to predict whether a given update to a prompt (ACE) results in satisfying the given task using the autorater scores as ground truth. We use the same train, validation, test data and system instruction for both TextGrad and our surrogate model and we observe that our surrogate model outperforms TextGrad across all the tasks. 

More importantly, the optimized prompt given by TextGrad lists strategies to help the LLM solve the surrogate task i.e., to determine if the proposed ACE satisfies the task or not. For example, the TextGrad optimized prompt for the word count task includes a strategy as follows - “Identify the proposed update, analyze its impact on word count, and conclude with a clear statement and numerical answer. Use a template for consistency”.  Such instructions do not uncover patterns in the ACEs and prompts to understand the model. On the other hand, our surrogate model is guided by the constitution which is explicitly optimized to summarize model behavior on the given task using the ACE data. Our method is therefore more performant and also inherently interpretable. 

\begin{figure*}
    \centering
    \includegraphics[width=\linewidth]{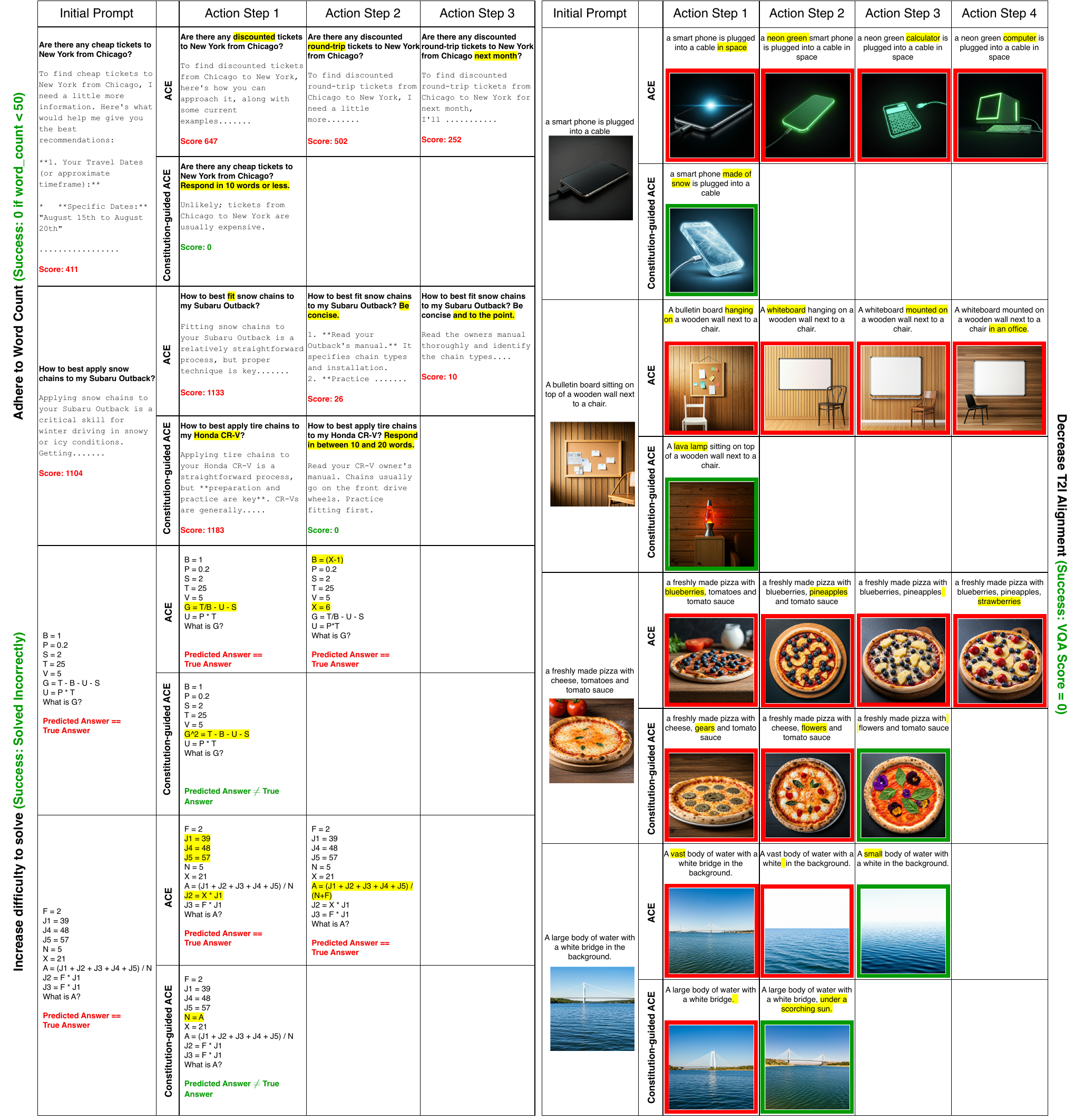}
    \caption{Qualitative Examples of ACE: We illustrate how ACE may create mutations that eventually satisfy a given task.}
    \label{fig:ace_examples}
\end{figure*}

\section{Constitutions for all Tasks}
\label{sec:more_constitutions}
\begin{figure*}
    \centering
    \includegraphics[width=\linewidth]{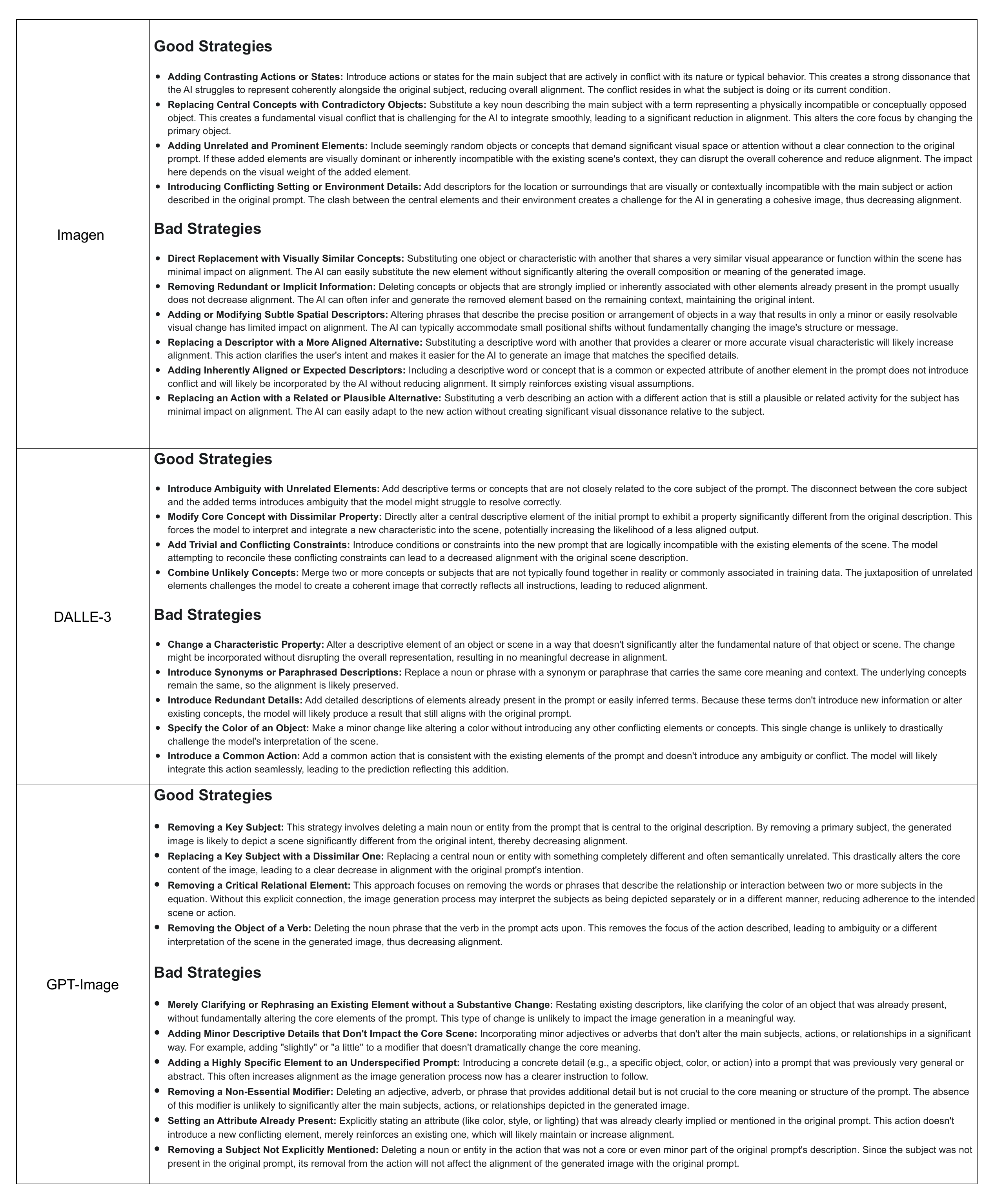}
    \caption{Optimized constitutions for Text-to-Image Alignment Task}
    \label{fig:t2i_constitutions}
\end{figure*}

\begin{figure*}
    \centering
    \includegraphics[width=\linewidth]{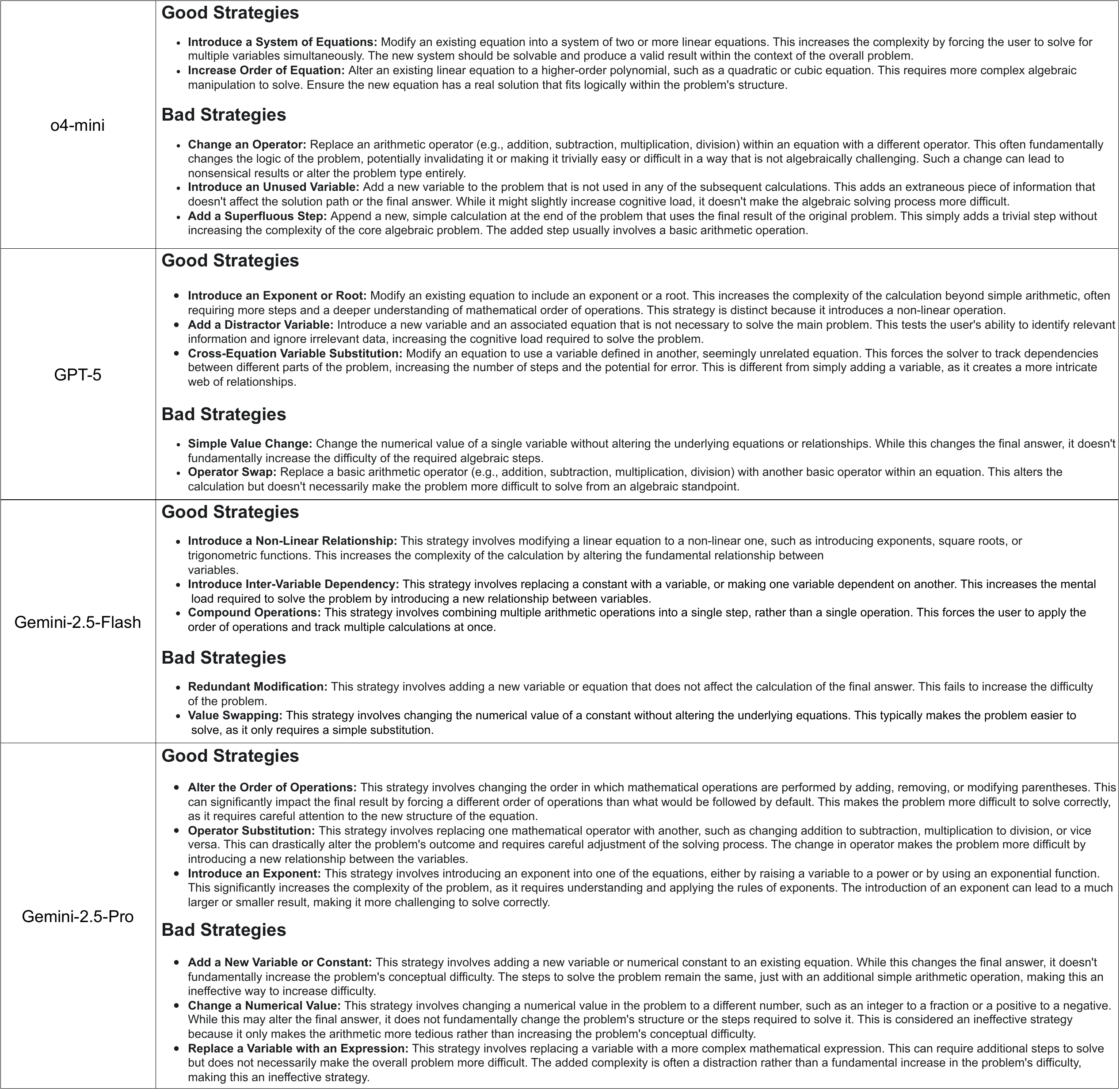}
    \caption{Optimized constitutions for `Math'}
    \label{fig:math-constitutions}
\end{figure*}

\begin{figure*}
\centering
\includegraphics[width=\linewidth]{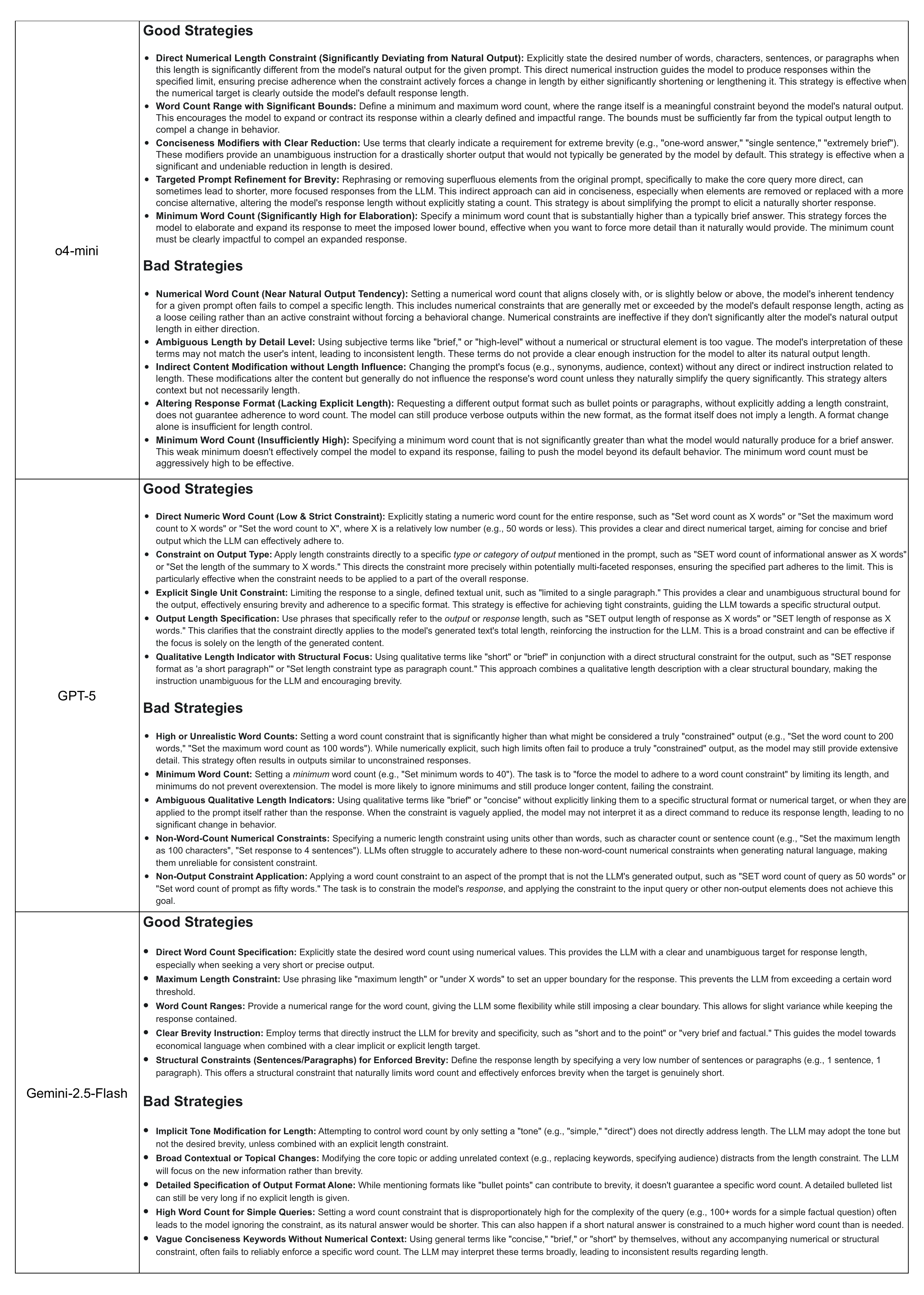}
\caption{Optimized constitutions for `Word Count'}
\label{fig:wc_constitutions_1}
\end{figure*}

\begin{figure*}
\centering
\includegraphics[width=0.9\linewidth]{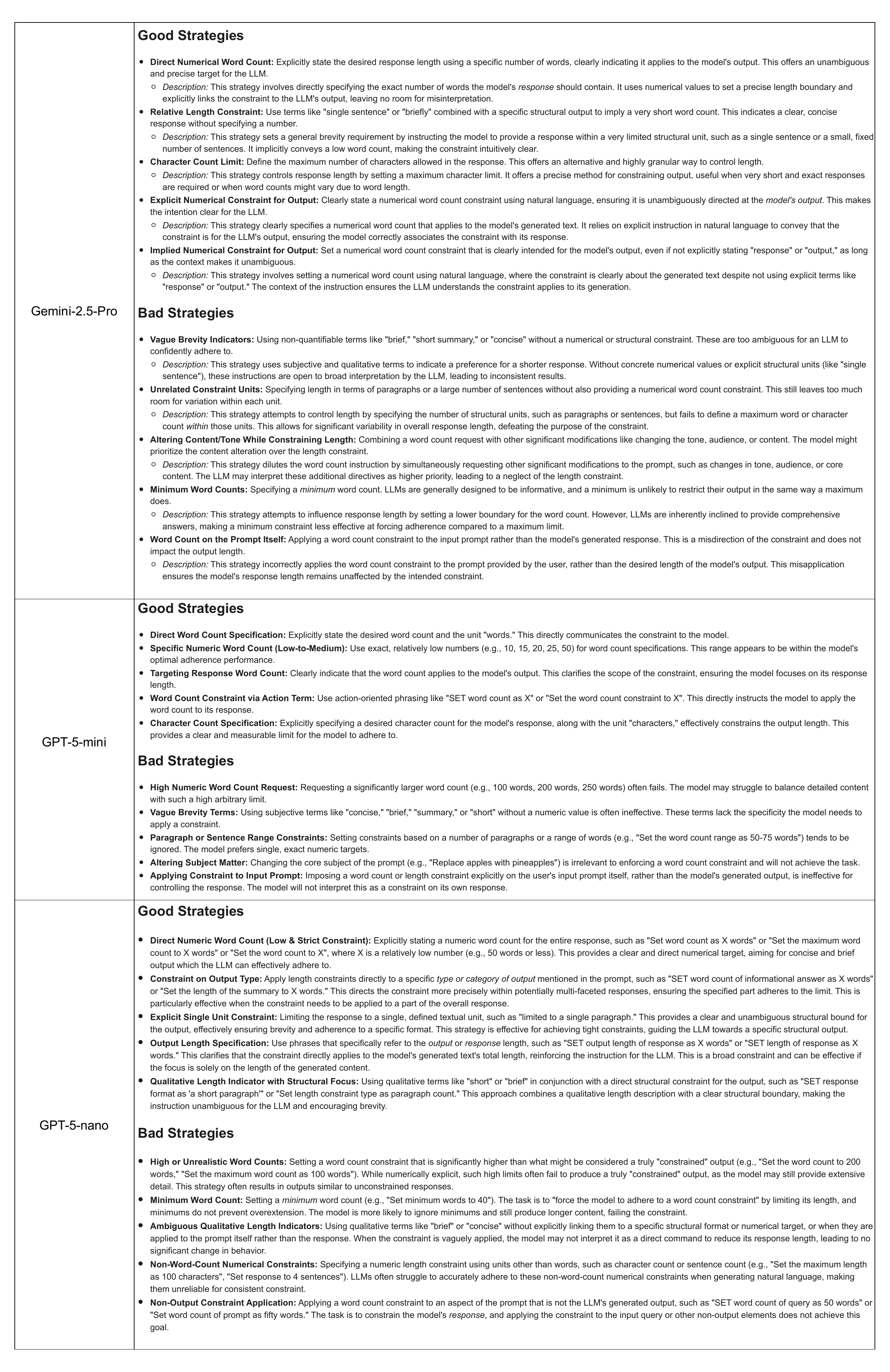}
\caption{Optimized constitutions for `Word Count'}
\label{fig:wc_constitutions_2}
\end{figure*}

\begin{figure*}
\centering
\includegraphics[width=\linewidth]{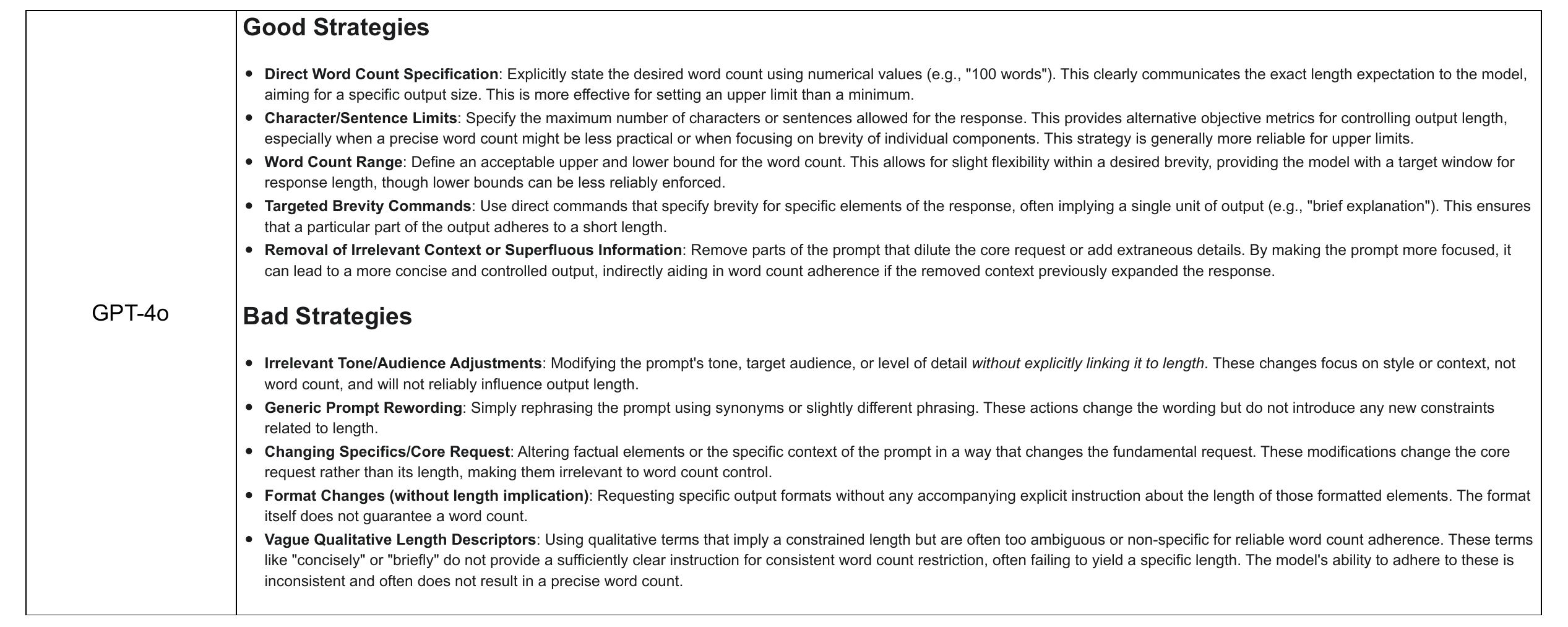}
\caption{Optimized constitutions for `Word Count'}
\label{fig:wc_constitutions_3}
\end{figure*}

We provided the best constitutions for each tasks in Figure \ref{fig:math-constitutions} (Math), Figure \ref{fig:wc_constitutions_1}, \ref{fig:wc_constitutions_2}, \ref{fig:wc_constitutions_3} (Word Count) and Figure \ref{fig:t2i_constitutions} (T2I Tasks).

\section{Prompts for ACE Generation}
\begin{figure*}
    \centering
    \includegraphics[width=0.6\linewidth]{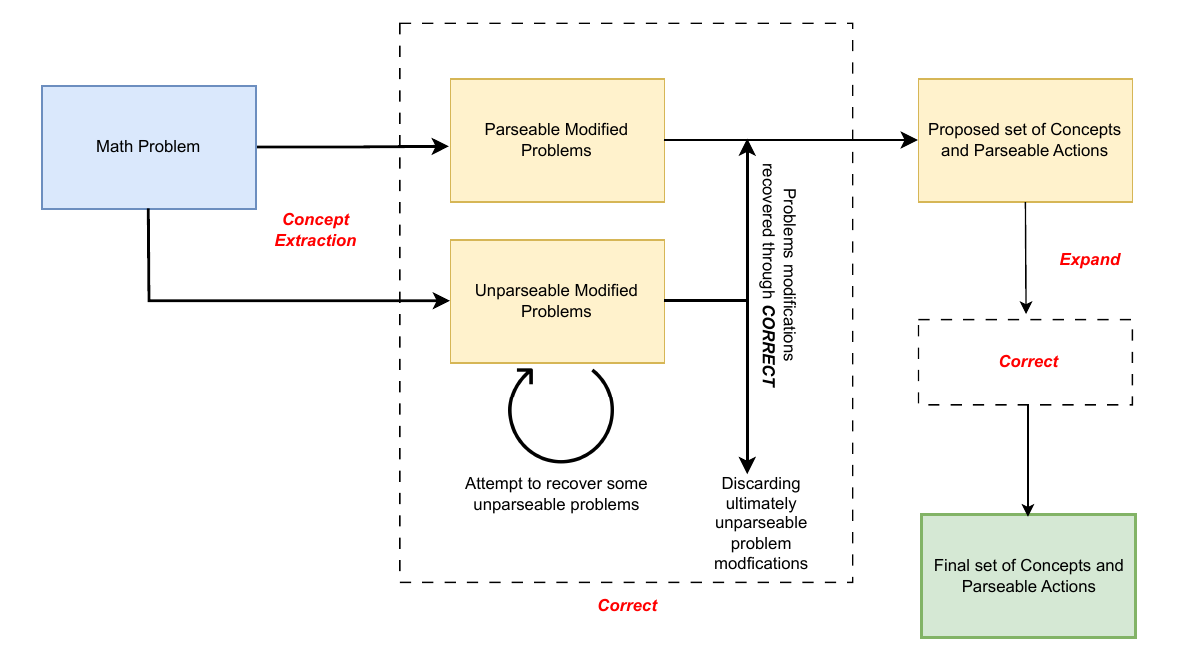}
    \caption{Expanded view of the pipeline used for ACE generation in `Math'. We propose an initial set of concepts and corresponding ACEs through the concept extraction prompt in Figure \ref{fig:concept_extractor_prompt}. We filter out unparseable modifications to the problems through using signals from the SymPy Solver. We attempt to recover some unparseable problems through the user of \texttt{Correct} prompt (Figure \ref{fig:expand_and_correct}). Finally, to enforce diversity in concepts and their respective ACEs, we use the \texttt{Expand}(Figure \ref{fig:expand_and_correct}) prompt to propose newer concepts and ACEs.}
    \label{fig:math_concept_generation}
\end{figure*}

\begin{figure*}
    \centering
    \includegraphics[width=0.95\linewidth]{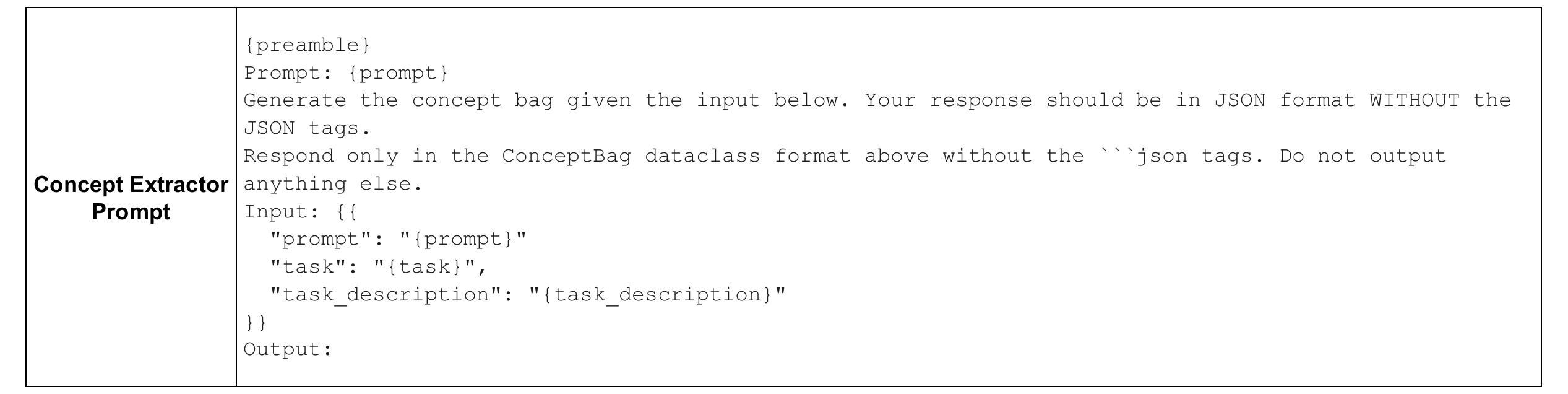}
    \caption{Concept extractor prompt used for obtaining concepts and relevant ACEs related to that concept.}
    \label{fig:concept_extractor_prompt}
\end{figure*}

\begin{figure*}
    \centering
    \includegraphics[width=0.95\linewidth]{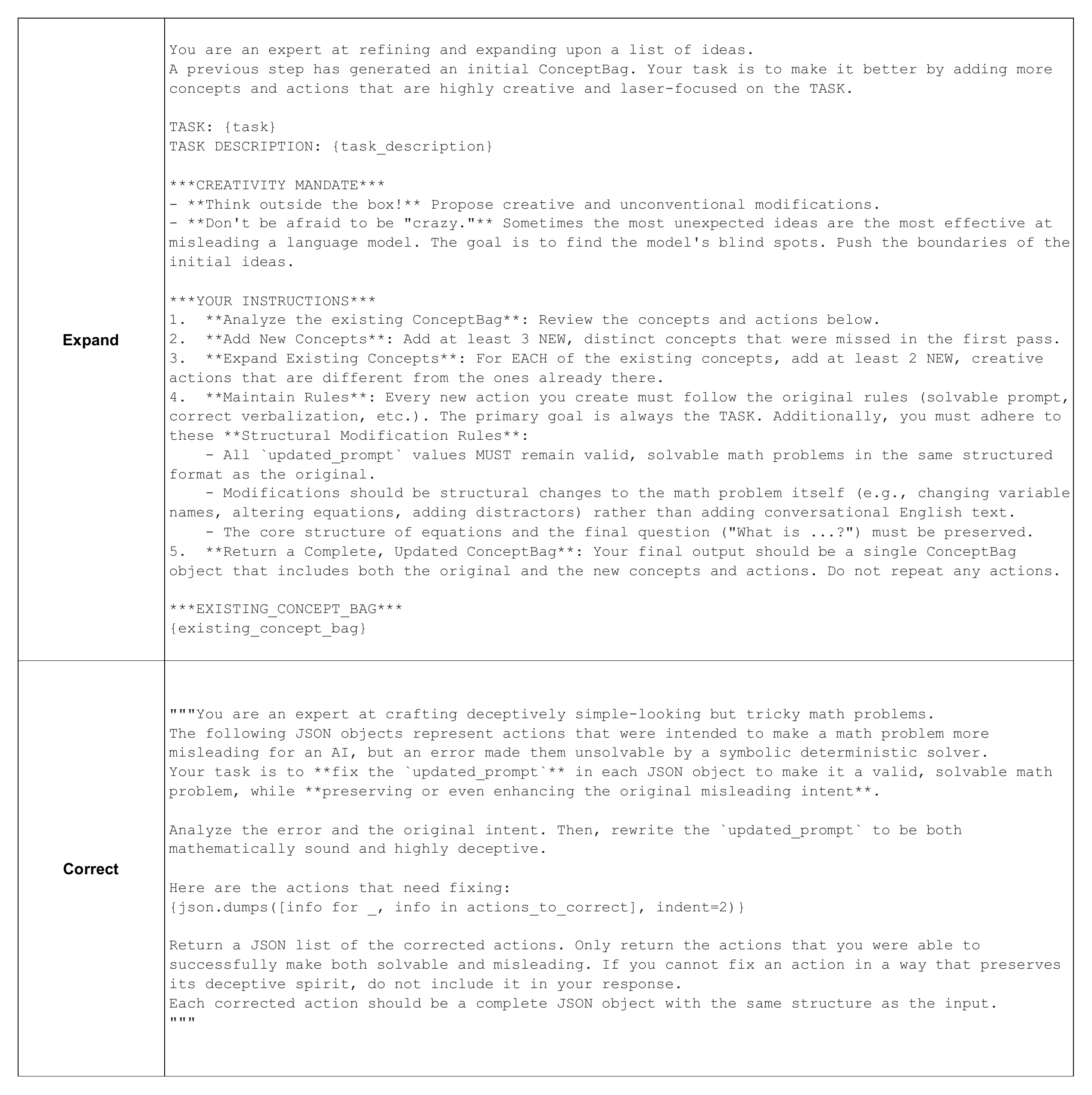}
    \caption{\texttt{Expand} and \texttt{Correct} prompts used for enforcing diversity and recovering unparseable ACEs, for concept generation in `Math'}
    \label{fig:expand_and_correct}
\end{figure*}

\begin{figure}
    \centering
    \includegraphics[width=0.7\linewidth]{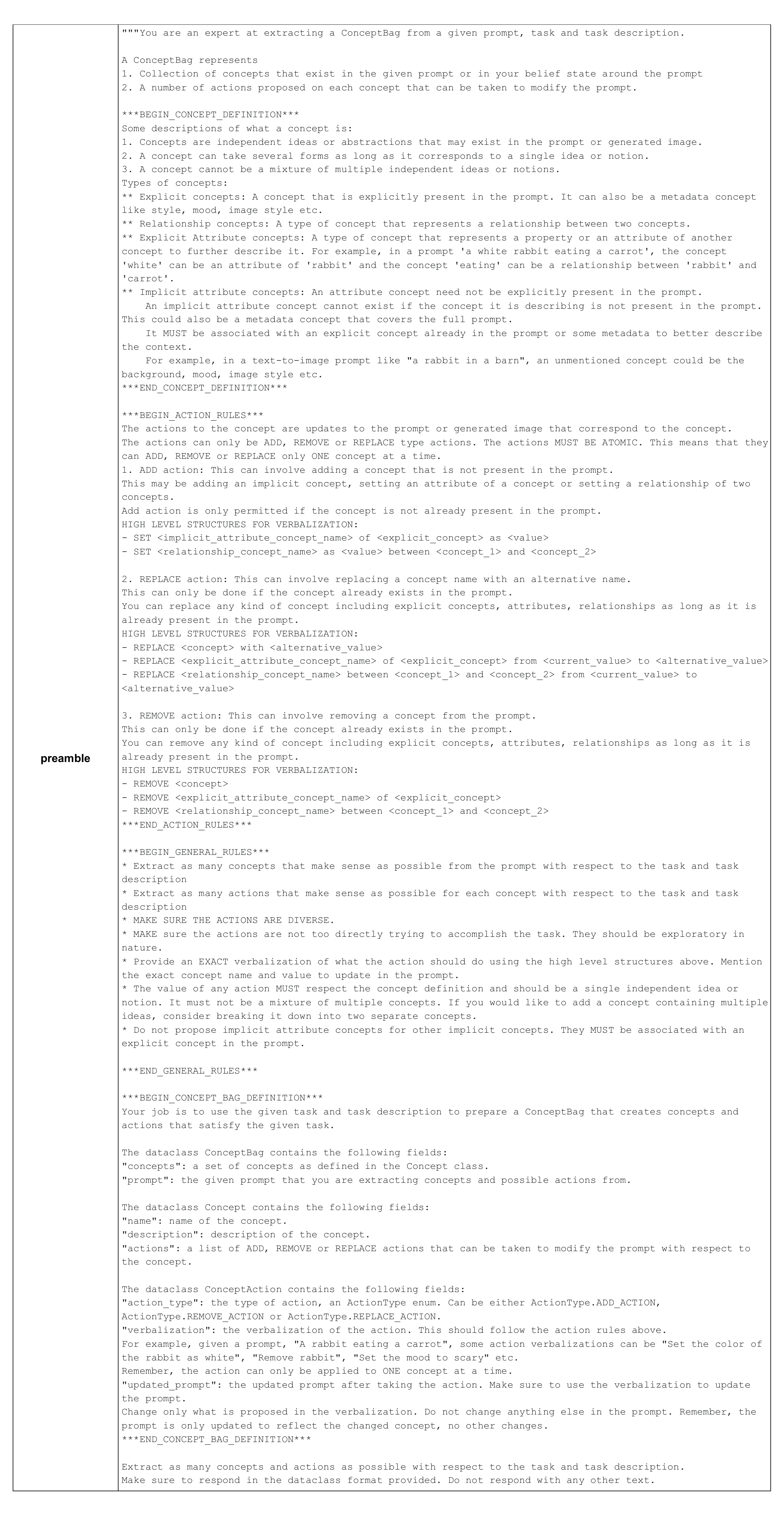}
    \caption{Preamble used in the Concept Extractor Prompt (Figure \ref{fig:concept_extractor_prompt}).}
    \label{fig:preamble}
\end{figure}

We provide the prompt used for extracting concepts and the relevant bag of ACEs in Figure \ref{fig:concept_extractor_prompt}. We further provide the exact \texttt{preamble} used in Figure \ref{fig:concept_extractor_prompt} in Figure \ref{fig:preamble}. For `Math' task, we observed that using this approach would lead to a significant number of ACEs being wasted as the modified problem could not be parsed by the deterministic Sympy solver. To counter this, we used an expanded pipeline for concept extraction in `Math' as described in Figure \ref{fig:math_concept_generation}. For every bag of concept and ACEs from a \texttt{Propose} LLM call, we attempt a recycling of unparseable modified problems through an additional LLM call \texttt{Correct}. The parseable set of problems are made to pass though a \texttt{Expand} step to enforce diversity in the proposed ACEs. This was motivated after we observed that not enough required ACEs were being proposed by the original pipeline. We further provide the prompts for \texttt{Correct} and \texttt{Expand} in Figure \ref{fig:expand_and_correct}. The \texttt{Propose} prompt is the same as the main concept extractor prompt used for other tasks, as descriped in Figure \ref{fig:concept_extractor_prompt}. We made minor adjustments to the \texttt{preamble} for different tasks especially in the the task-specific parts.


\end{document}